\documentclass[12pt]{article}
\usepackage{template}
\usepackage{xcolor}
\usepackage{algorithm}
\usepackage{algpseudocode}

\begin{document}

\thispagestyle{empty} 
\baselineskip=28pt

\begin{center}
    {\LARGE{\bf Echo State Networks for Spatio-Temporal Area-Level Data}}
\end{center}

\baselineskip=12pt

\vskip 2mm
\begin{center}
Zhenhua Wang\footnote{(\baselineskip=10pt to whom correspondence should be
  addressed) Department of Statistics, University of Missouri, 146 Middlebush Hall,
  Columbia, MO 65211-6100, zhenhua.wang@missouri.edu}\,
Scott H. Holan\footnote{\baselineskip=10pt Department of Statistics, University of
  Missouri, 146 Middlebush Hall, Columbia, MO 65211-6100,
  holans@missouri.edu}\,\footnote{\baselineskip=10pt Office of the Associate Director for Research and Methodology, U.S. Census Bureau, 4600 Silver
  Hill Road, Washington, D.C. 20233-9100, scott.holan@census.gov}\, and 
Christopher K. Wikle \footnote{\baselineskip=10pt Department of Statistics, University of
  Missouri, 146 Middlebush Hall, Columbia, MO 65211-6100,
  wiklec@missouri.edu}

\end{center}

\vskip 4mm

\begin{center}
    {\bf Abstract}
\end{center}

\baselineskip=12pt 

Spatio-temporal area-level datasets play a critical role in official statistics, providing valuable insights for policy-making and regional planning. Accurate modeling and forecasting of these datasets can be extremely useful for policymakers to develop informed strategies for future planning. Echo State Networks (ESNs) are efficient methods for capturing nonlinear temporal dynamics and generating forecasts. However, ESNs lack a direct mechanism to account for the neighborhood structure inherent in area-level data. Ignoring these spatial relationships can significantly compromise the accuracy and utility of forecasts. In this paper, we incorporate approximate graph spectral filters at the input stage of the ESN, thereby improving forecast accuracy while preserving the model's computational efficiency during training. We demonstrate the effectiveness of our approach using Eurostat's tourism occupancy dataset and show how it can support more informed decision-making in policy and planning contexts.

\vskip 4mm

\par\vfill\noindent

\clearpage
\pagenumbering{arabic}
\baselineskip=24pt

\section{Introduction}
\label{sec:intro}
Spatio-temporal area-level datasets are fundamental in official statistics, providing essential data for policy-making and regional planning. For instance, the Bureau of Labor Statistics (BLS) produces the Quarterly Census of Employment and Wages (QCEW) which can be used to analyze business dynamics across the U.S., the Census Bureau produces the Longitudinal Employer-Household Dynamics (LEHD) which can be used to study the flow of jobs and workforces, and the Centers for Disease Control and Prevention (CDC) disseminates numerous spatio-temporal datasets that can be used to track the spread of diseases. Accurately modeling these datasets and generating reliable forecasts can be critical for policymakers to craft informed strategies for future planning. These datasets often exhibit complex temporal dynamics and strong spatial structures, making it crucial to incorporate both temporal and spatial components into the models. Neglecting either aspect can result in inaccurate estimates and potentially misguided policy decisions for data users\citep{cressie2011statistics, cressie2015statistics, porter2014spatial, porter2015small}. 

A widely used method for handling such datasets is Dynamic Linear Models (DLM), a versatile class of models designed for nonstationary time series that can incorporate various temporal components, such as trend and seasonality \citep{west2006bayesian, prado2010time, dlm_book, dlm_r}. However, DLMs are primarily suited for short-lead forecasting. A significant drawback is that they often require substantial computational resources and may encounter numerical issues when applied to large datasets or complex geometries \citep{mcdermott2017ensemble, mcdermott2019deep}. Additionally, the modeling process involves making subjective decisions about which state components to include, adding complexity to the model specification.

Recent developments in machine learning and deep learning bring new opportunities in modeling statistical time series. For example, \citet{eliwa2024optimal} uses the adaptive network-based fuzzy inference system to predict gasoline price. \citet{hassan2024optimizing} uses language models to automatically predict diseases from their symptoms. Among all machine learning models, Recurrent Neural Networks (RNN) are highly flexible models capable of capturing complex dynamic patterns through recursive connections between and within neurons \citep{rumelhart1986learning, hochreiter1997long}. The Echo State Network (ESN) is a specialized type of RNN that efficiently reduces computational requirements by randomly sampling most parameters rather than learning them from data. ESNs have demonstrated success in various statistical time series tasks \citep{jaeger2002adaptive, lukovsevivcius2012practical}. For instance, \citet{mcdermott2017ensemble, bonas2024calibrated} employed ESNs to forecast Pacific sea surface temperature, \citet{mcdermott2019deep} applied ESNs for soil moisture forecasting, \citet{yoo2023using} utilized ESNs to model wildfire front propagation, and \citet{grieshop2024echo} embedded ESNs within a Bayesian hierarchical model to forecast the spread of raccoon rabies. \citet{wang2025hierarchical} extended ESNs to accommodate count data and applied them to forecast the number of graduate students across universities in the United States. In such applications, spatio-temporal data is often treated as a multivariate time series, with all spatial locations considered as multiple responses, or transformed using empirical orthogonal functions (EOF), with ESN applied to model the EOF coefficients. This approach commonly involves selecting a number of EOFs to retain, which can be determined using the elbow method, setting a threshold for the percentage of variance explained, or by comparing the singular values with those obtained from equivalent uncorrelated data \citep{hastie2009elements}. However, this general approach may result in the loss of essential information during the dimension reduction process, potentially leading to reduced forecasting performance.


To address this, we aim to capture the neighborhood structure through an embedding similar to Graph Convolutional Networks (GCN). GCNs are specialized convolutional layers in deep learning designed to capture complex features within graph data \citep{kipf2016semi}. They are widely regarded as the state-of-the-art method for learning graph representations. Since area-level data can essentially be viewed as an undirected graph, GCN could be well-suited for this task. GCNs operate by averaging the hidden representations of each node with those of its neighbors, and optimizing the target function using gradient-based methods. Although GCNs handle graph features effectively, a key limitation is the significant resource and time requirements for training, as GCNs require the entire graph to be processed in every batch. Various approaches have been proposed to alleviate this issue. For instance, \citet{wu2019simplifying} show that the nonlinear activation in GCNs is not essential, as most of the benefits arise from the local averaging in the GCN layer. They further propose the Simplified Graph Convolutional Network (SGC), which employs a linear activation combined with local averaging among neighbors. Additionally, several authors have explored randomization-based methods to reduce the computational overhead in GCNs. For instance, \citet{huang2022graph} demonstrate that employing random weights in GCNs can significantly enhance efficiency while maintaining strong performance in node classification tasks. \citet{liu2021random} combine GCNs with kernel learning using random features, showing promising results in both node classification and clustering tasks.

Building on this idea, we treat the task of handling spatial features as a problem of learning good representations from data \citep{bengio2013representation}. Specifically, we draw inspiration from the expanded representation concept proposed by \citet{sutton2014online}, which involves a two-layer model: the first layer expands the original input into a high-dimensional feature space, while the second layer maps this representation to the output. Classic models such as Rosenblatt’s perceptron and modern deep learning frameworks can all be seen as instances of expanded representation learning. Notably, \citet{gallant1987random} and \citet{sutton2014online} advocate the use of a random fixed layer for input expansion, a technique known as random representation. This method efficiently handles high-dimensional inputs and enables faster computation, making it well-suited for generating input embeddings for ESN.

In this paper, we propose a novel approach for capturing spatio-temporal dynamics often found in areal data and apply the approach to data disseminated by Eurostat. We employ an ensemble ESN with a specialized input embedding designed for areal data, integrating concepts from Random Representation and SGC layer. We refer to the ESN with areal random representation at the input stage as AESN. Our spatial input embedding leverages the local averaging properties of the SGC, and uses the fixed randomly sampled parameters to effectively capture spatial features. By addressing spatial dependencies at the input stage, this approach significantly enhances ESN performance while maintaining its computational efficiency. 

We demonstrate the performance of the proposed method using Eurostat's tourism occupancy data. Tourism plays a crucial role in shaping regional economies, and tourism occupancy is a valuable indicator for the popularity of the tourism industry in Europe. It is essential to understand the spatio-temporal patterns of tourism occupancy and accurately forecast them into the future. However, modeling this data poses significant challenges due to its inherent complexity. As an area-level survey dataset, tourism occupancy data exhibits intricate spatial patterns within the European NUTS \footnote{\baselineskip=10pt Nomenclature of Territorial Units for Statistics (NUTS) is the official classification of subregions used by Eurostat.} level 2 regions, coupled with complex temporal dynamics. These characteristics make the proposed method suitable for capturing the underlying patterns.

The paper is organized as follows. Section~\ref{sec:background} reviews the concepts of ESN and GCN. In Section~\ref{sec:method}, we describe our areal input embedding, along with its uncertainty quantification and hyperparameter selection. Section~\ref{sec:analysis} provides an illustration of forecasting performance on the Eurostat's night spent at tourism accommodations dataset. Finally, we provide a brief summary in Section~\ref{sec:discussion}.

\section{Background}
\label{sec:background}
\subsection{Echo State Network}
As described in \citet{mcdermott2017ensemble}, the fundamental architecture of an ESN is described as follows
\begin{align}
\text{output layer}: \qquad &\boldsymbol{Y}_t = \boldsymbol{W}_{out} \boldsymbol{h}_t + \boldsymbol{\epsilon}_t, \boldsymbol{\epsilon}_t \sim N(\boldsymbol{0}, \sigma^2 \boldsymbol{I}), \nonumber\\
\text{hidden layer}: \qquad &\boldsymbol{h}_t = (1 - \alpha) \boldsymbol{h}_{t-1} + \alpha \widetilde{\boldsymbol{h}}_t,\nonumber\\
&\widetilde{\boldsymbol{h}}_t = g_h\left(\frac{\nu}{|\lambda_w|}\boldsymbol{W}_{res}\boldsymbol{h}_{t-1} + \boldsymbol{W}_{in} \widetilde{\boldsymbol{x}}_t\right), \nonumber\\
\text{parameters}: \qquad &\boldsymbol{W}_{res} = [w_{i, l}]_{i, l}: w_{i, l} = \gamma_{i, l}^{res} \text{Unif}(-a_{res}, a_{res}) + (1 - \gamma_{i, l}^{res})\delta_0, \label{eq:esn_wres}\\
&\boldsymbol{W}_{in} = [w_{i, j}]_{i, j}: w_{i, j} = \gamma_{i, j}^{in} \text{Unif}(-a_{in}, a_{in}) + (1 - \gamma_{i, j}^{in})\delta_0, \label{eq:esn_win}\\
&\gamma_{i, l}^{res} \sim \text{Bern}(\pi_{res}), \nonumber\\
&\gamma_{i, j}^{in} \sim \text{Bern}(\pi_{in}),\nonumber
\end{align}
where $\boldsymbol{Y}_t \in \mathbb{R}^{n_y}$, $\boldsymbol{h}_t \in \mathbb{R}^{n_h}$, and $\tilde{\boldsymbol{x}_t} \in \mathbb{R}^{n_x}$ represent the output vector, the hidden state vector, and the input vector at time $t$, respectively. The matrices $\boldsymbol{W}_{out} \in \mathbb{R}^{n_y \times n_h}$, $\boldsymbol{W}_{res} \in \mathbb{R}^{n_h \times n_h}$, and $\boldsymbol{W}_{in} \in \mathbb{R}^{n_h \times n_x}$ denote the output weights, the hidden weights, and the input weights at time $t$, respectively. $\delta_0$ denotes the Dirac function centered at zero, $\alpha$ is the leaking rate from the previous reservoir state, $\nu$ is the scaling factor for $\boldsymbol{W}_{res}$, and $\lambda_w$ is the largest eigenvalue of $\boldsymbol{W}_{res}$.

The hidden layer, often known as the reservoir computing layer, plays a crucial role in modeling temporal dynamics. The reservoir itself is constructed using the randomly sampled weight matrix, $\boldsymbol{W}_{res}$, which is scaled by its spectral radius, $|\lambda_w|$. This process creates a large pool of nonlinear transformations of the input signals, combining the scaled input data from the current time with outputs from the previous reservoir states. These transformations, or hidden states, are designed to capture the underlying dynamics of the time series. A fundamental aspect of ESNs is their reliance on the echo state property (ESP), which states that the reservoir state is driven by the past input sequence, rather than its initial state \citep{jaeger2002adaptive, lukovsevivcius2012practical}. In practice, the ESP is typically maintained by constraining the spectral radius of the reservoir weights to be less than one \citep{jaeger2002adaptive, lukovsevivcius2012practical}.

Once the hidden states are generated, the output weights $\boldsymbol{W}_{out}$, also called readout weights, are trained to map these states to the desired output. Notably, the only parameter that requires training in this approach is the readout weights, which can be efficiently estimated using simple ridge regression. The loss function for this estimation is given by $L = \frac{1}{N} \sum_{t=1}^{T} \left( (\boldsymbol{Y}_t - \boldsymbol{W}_{out} \boldsymbol{h}_t) (\boldsymbol{Y}_t - \boldsymbol{W}_{out} \boldsymbol{h}_t)' \right) + \tau \sum_{j=1}^{n_y} \| \boldsymbol{W}_{out}^2 \|_2$, where $\tau$ is the penalty term. As a result, reservoir computing only requires a minimal amount of computation resources.

To ensure optimal performance, the hyperparameters of the ESN must be fine-tuned. Specifically, we need to determine the best values for the reservoir size, $\boldsymbol{W}_{res}$, denoted as $n_h$, the scaling factor of $\boldsymbol{W}_{res}$, $\nu$, the distribution parameters of the reservoir and input weights, the leaking rate $\alpha$ from the previous reservoir state, and the penalty coefficient in ridge regression, $\tau$. In practice, particularly in spatio-temporal modeling applications, the most influential hyperparameters are $\nu$, $n_h$, and $\tau$ \citep{mcdermott2017ensemble, yoo2023using}.

\subsection{Graph Convolutional Network}
\label{sec:gcn}
\citet{kipf2016semi} introduced the GCN to learn feature representations for each node in a graph. Given a graph $\mathcal{G}$ with $n$ nodes, each having $p$ features, and a proximity matrix $\boldsymbol{A} \in \mathbb{R}^{n \times n}$, let $\boldsymbol{H}_0 = \boldsymbol{X} \in \mathbb{R}^{n \times p}$ represent the input feature matrix. For $i > 1$, let $\boldsymbol{H}_{i} \in \mathbb{R}^{n \times h_i}$ denote the feature matrix learned at the $i$th graph convolutional layer. $\boldsymbol{H}_{i}$ is updated as 
\begin{align}
    \boldsymbol{H}_{i} &= \boldsymbol{S} \boldsymbol{H}_{i-1} \boldsymbol{\Theta}_{i-1}, \label{eq:GCN}\\
    \boldsymbol{S} &= \widetilde{\boldsymbol{D}}^{-\frac{1}{2}} \widetilde{\boldsymbol{A}} \widetilde{\boldsymbol{D}}^{-\frac{1}{2}},
\end{align}
where $\boldsymbol{\Theta}_{i-1} \in \mathbb{R}^{h_{i-1} \times h_i}$ is the weight matrix at the (i-1)th layer, $\widetilde{\boldsymbol{A}} = \boldsymbol{A} + \boldsymbol{I} \in \mathbb{R}^{n \times n}$, which denotes the proximity matrix with self-loops, and $\widetilde{\boldsymbol{D}} \in \mathbb{R}^{n \times n}$ is the diagonal degree matrix with entries given by the row sums of $\widetilde{\boldsymbol{A}}$. This definition of the proximity matrix is beneficial because it allows for handling isolated nodes, where the node's features consist solely of its own attributes rather than incorporating features from neighboring nodes.

\citet{kipf2016semi} further show that the update rule of GCNs can be viewed as a first-order approximation of localized spectral filters on graphs. Let $\boldsymbol{L} \in \mathbb{R}^{n \times n}$ denotes the normalized graph Laplacian of $\mathcal{G}$, $\boldsymbol{U} \in \mathbb{R}^{n \times n}$ be the eigenvectors of $\boldsymbol{L}$, and $\boldsymbol{\Lambda} \in \mathbb{R}^{n \times n}$ be the diagonal matrix of eigenvalues of $\boldsymbol{L}$. We can write $\boldsymbol{L}$ as 
\[
\boldsymbol{L} = \boldsymbol{I} - \boldsymbol{D}^{-1/2} \boldsymbol{A} \boldsymbol{D}^{-1/2} = \boldsymbol{U} \boldsymbol{\Lambda} \boldsymbol{U}',
\]
where $\boldsymbol{A}$ is defined as in GCN and $\boldsymbol{D}$ is the corresponding degree matrix. The spectral convolution operation on graph can be defined as
\begin{align}
\boldsymbol{g}_{\theta} * \boldsymbol{x} = \boldsymbol{U} \boldsymbol{g}_{\theta} \boldsymbol{U}' \boldsymbol{x},
\label{eq:graph_conv}
\end{align}
where $\boldsymbol{x}$ is the input signal on the nodes, and $\boldsymbol{g}_{\theta}$ is a spectral filter parameterized by $\theta$ in the Fourier domain.
\citet{kipf2016semi} approximate $\boldsymbol{g}_{\theta}$ using first-order Chebyshev polynomials $\boldsymbol{g}_{\theta} \approx \theta_0 + \theta_1 \boldsymbol{\Lambda}$ \citep{hammond2011wavelets}. Thus, (\ref{eq:graph_conv}) can be effectively rewritten as
\begin{align*}
\boldsymbol{g}_{\theta} * \boldsymbol{x} \approx 2\theta_0 \boldsymbol{x} - \theta_1 \boldsymbol{D}^{-1/2} \boldsymbol{A} \boldsymbol{D}^{-1/2} \boldsymbol{x}.
\end{align*}
To address overfitting and reduce the number of parameters, \citet{hammond2011wavelets} further set $\theta = 2\theta_0 = -\theta_1$, and allow the parameter $\theta$ to be shared across the entire graph. Additionally, to avoid numerical instabilities, they redefine the adjacency and degree matrices as $\widetilde{\boldsymbol{A}} = \boldsymbol{A} + \boldsymbol{I}$ and $\widetilde{\boldsymbol{D}}_{ii} = \sum_j \widetilde{\boldsymbol{A}}_{ij}$, respectively. We thus obtain the final form of the GCN updating rule as shown in (\ref{eq:GCN}).

\citet{kipf2016semi} consider applying a nonlinear activation function to the output of (\ref{eq:GCN}) to make GCNs more flexible. However, \citet{wu2019simplifying} show that the ability for feature extraction on graphs is due to the local averaging introduced by (\ref{eq:GCN}), and that the nonlinear activation is not necessary. They denote the GCN with a linear activation function as the Simple Graph Convolution (SGC) and demonstrate that it corresponds to a fixed low-pass filter on the graph spectral domain.



\section{Method}
\label{sec:method}
\subsection{Areal Random Representation}
In addition to the reservoir, the input signal is another important component of the ESN. It can include covariates that account for confounding effects not captured by the time series alone. When additional covariate variables are unavailable, lagged values are typically used as input signals, as their trajectories describe the temporal dynamics of the series \citep{mcdermott2017ensemble, mcdermott2019deep, yoo2023using}. In spatial statistics, spatial input embedding is widely used to capture the spatial structure found in the data. For example, \citet{daw2023reds} use random Fourier features to approximate the stationary Gaussian process kernel. However, there is no inherent way of incorporating neighborhood structures into the ESN. The common approach for capturing spatial features in ESNs involves modeling the EOFs coefficients asscociated with a spatio-temporal series \citep{mcdermott2017ensemble, mcdermott2019deep, yoo2023using}. This approach, however, requires deciding how many EOFs to include and may lead to a potential loss in forecasting performance, as demonstrated in our data analysis.

As mentioned in Section \ref{sec:intro}, we leverage random representations to capture spatial dynamics at the input stage \citep{bengio2013representation, gallant1987random, sutton2014online}. Building on the approach of \citet{gallant1987random}, we represent spatial features found in areal data through a dense embedding of input signals, using a structure inspired by SGC layers \citep{wu2019simplifying}. To motivate our embedding method, we can revisit (\ref{eq:GCN}) in the SGC layer. Let $\boldsymbol{X} \in \mathbb{R}^{n_s \times n_x}$ denote the input signals, $\boldsymbol{A} \in \mathbb{R}^{n_s \times n_s}$ the proximity matrix, and $\widetilde{\boldsymbol{A}} = \boldsymbol{I} + \boldsymbol{A}$ the proximity matrix with self-loops. We further let $\widetilde{\boldsymbol{D}}$ represent the degree matrix of $\widetilde{\boldsymbol{A}}$, and $\boldsymbol{S} = \widetilde{\boldsymbol{D}}^{-1/2} \widetilde{\boldsymbol{A}} \widetilde{\boldsymbol{D}}^{-1/2}$ the symmetric normalized adjacency matrix. With some simple linear algebra, (\ref{eq:GCN}) can be rewritten as
\begin{equation*}
    \boldsymbol{z}^{(k)}(l) = \sum_{l' \in N(l)} S(l,l') \boldsymbol{X}(l)\boldsymbol{W}^{(k)},
\end{equation*}
where $S(l, l')$ denotes the entry of $\boldsymbol{S}$ at the $l$-th row and $l'$-th column, $\boldsymbol{z}^{(k)}(l)$ is the $k$-th embeded feature at location $l$, and $\boldsymbol{W}^{(k)}$ is the $k$-th kernel of dimension $n_x$. We note that $\boldsymbol{W}^{(k)}$ is shared across all locations, which may be less effective at capturing graph features \citep{liu2020comprehensive}. To enhance its capacity, we construct the random representation using spatially varying kernels, which we refer to as the areal random representation. The areal random representation is defined as
\begin{eqnarray}
\boldsymbol{z}^{(k)}(l) &=& \sum_{l' \in \mathcal{N}(l)} S(l,l')\boldsymbol{X}(l) \boldsymbol{U}^{(k)}(l'),\label{eq:embed} \\
\boldsymbol{U}^{(k)} &\sim& \text{Unif}(-a_u, a_u),\nonumber
\end{eqnarray}
where $\mathcal{N}(l)$ represents the neighborhood of location $l$ defined by $\widetilde{\boldsymbol{A}}$. For $k = 1, \dots, K$, $\boldsymbol{z}^{(k)}(l)$ is the $k$-th embeded feature at location $l$, and $\boldsymbol{U}^{(k)}(l')$ are $n_x$ randomly sampled weights at location $l'$. To further expand the capacity of representation, we generate \( K \) different copies of locally averaged random signals for each \( \boldsymbol{U}^{(k)} \). The resulting embedding matrix \( \mathbf{Z} \) is the collection of \( \{ \mathbf{z}^{(1)}, \dots, \mathbf{z}^{(K)} \} \), with dimensions \( n_s \times K \). Here, we note that the parameter $K$ closely resembles the number of neurons or feature maps in a neural network layer. These weights can also be interpreted as the approximated localized spectral filters with different kernels, enabling the model to capture a rich set of local features. Hence, we embed the input signals at each location from dimension \( n_x \) to \( K \). We also note that these weights are sampled without access to the data, which prevents the embedding layers from learning noise. To control overfitting for the entire AESN model, regularization should be applied to the ridge regression at the output layer. Lastly, we note that (\ref{eq:embed}) can be vectorized as
\begin{eqnarray}
    \boldsymbol{z}^{(k)} = \boldsymbol{S} \sum_{j=1}^{n_x} (\boldsymbol{U}^{(k)} \odot \boldsymbol{X})_j,\label{eq:embed_vec}
\end{eqnarray}
where $\odot$ is the the Hadamard product. To simplify notation, we use $(\cdots)_j$ to denote the $j$-th column of the matrix within the parentheses. In (\ref{eq:embed_vec}), we first compute the column sums of $(\boldsymbol{U}^{(k)} \odot \boldsymbol{X})$, then apply a matrix product with $\boldsymbol{S}$ to filter out relevant neighbors. 

To integrate these embedded signals into the ESN, we apply the areal random representation to the input signals $\widetilde{\boldsymbol{x}}_t$ at each time step, resulting in the embedded signals $\boldsymbol{z}_t^{(k)}$, as shown in (\ref{eq:ARER}). The vectorized signals $\widetilde{\boldsymbol{z}}_t$ are then used as the input to the ESN, as in (\ref{eq:ARER_vec}). This implies that when a signal at any location is activated in the reservoir stage, all signals in its neighborhood are also activated. AESN can be defined as, 
\begin{align}
\boldsymbol{Y}_t &= \boldsymbol{W}_{out} \boldsymbol{h}_t + \boldsymbol{\epsilon}_t, \boldsymbol{\epsilon}_t \sim N(\boldsymbol{0}, \sigma^2 \boldsymbol{I}) \nonumber\\
\boldsymbol{h}_t &= (1 - \alpha) \boldsymbol{h}_{t-1} + \alpha \widetilde{\boldsymbol{h}}_t\label{eq:aesn_h1}\\
\widetilde{\boldsymbol{h}}_t &= g_h(\frac{\nu}{|\lambda_w|}\boldsymbol{W}_{res}\boldsymbol{h}_{t-1} + \boldsymbol{W}_{in} \widetilde{\boldsymbol{z}}_t)\label{eq:aesn_h2}\\
\widetilde{\boldsymbol{z}}_t &= \text{vec}(\boldsymbol{Z}_t) \label{eq:ARER_vec}\\
\boldsymbol{z}_t^{(k)} &= \boldsymbol{S} \sum_{j = 1}^{n_x}(\boldsymbol{U}^{(k)} \odot \widetilde{\boldsymbol{x}}_t)_j  \label{eq:ARER}\\
\boldsymbol{U}^{(k)} &\sim \text{Unif}(-a_u, a_u),\label{eq:u_embed}
\end{align}
where $\boldsymbol{Z}_t \in \mathbb{R}^{n_s \times K}$ is the embedding matrix whose $k$th column is $\boldsymbol{z}_t^{(k)}$, and $\text{vec}(\cdot)$ is the vectorization operator. Here, it is worth emphasizing that capturing spatial dynamics at the input stage offers two key advantages. First, manipulating the input stage preserves the all desirable properties of the ESN, such as the ESP. Second, handling spatial dynamics at the input stage introduces minimal computational overhead, as it avoids being recursively computed at every time step.

\subsection{Hyperparameter selection}
\label{sec:tune}
Hyperparameter selection plays a critical role in the performance of an ESN, as it significantly impacts the flexibility, stability, and generalization of the model. The most common approaches for hyperparameter selection are grid search \citep{mcdermott2017ensemble, mcdermott2019deep, lukovsevivcius2012practical} and random search \citep{yoo2023using, bergstra2012random}. Another approach involves combining results from ESNs with different hyperparameter sets, which can be viewed as model averaging \citep{grieshop2024echo}. In this work, we opted for the random search method, as it is more efficient in exploring the possible hyperparameter space.

To tune the AESN, we first determine the distribution and its sparsity parameter, following the guidelines of \citet{lukovsevivcius2012practical} and \citet{mcdermott2017ensemble}. Specifically, we sample $\boldsymbol{W}_{\text{res}}$ from a mixture of the Dirac delta function $\delta_0$ and a uniform distribution, setting \{$a_{\text{in}}$,$a_{\text{res}}$\} to 0.1. In accordance with \citet{lukovsevivcius2012practical}, we make $\boldsymbol{W}_{\text{in}}$ and $\boldsymbol{U}^{(k)}$ dense by sampling from a uniform distribution. We then apply random search to tune $a_u$, $\nu$, $n_h$, $k$, and the penalty parameter $\tau$ in ridge regression. Specifically, we split the validation data from the training data. In each iteration of the random search, we sample the hyperparameters from log-uniform distributions. The search ranges are $(15, 300)$ for $K$ and $n_h$, $(0.1, 1)$ for $a_u$, $(0.05, 1)$ for $\nu$, and $(0.001, 0.1)$ for $\tau$. Here, the log-uniform distribution is used for its efficiency in exploring values that vary across different orders of magnitude \citep{bengio2013representation}. For integer parameters, we round the sampled values to the nearest integer. The optimal set of parameters is selected based on the best forecast performance on the validation set. 


\subsection{Ensemble AESN}
The uncertainty quantification for the proposed model follows the ensemble method outlined in \citet{mcdermott2017ensemble, mcdermott2019deep}, incorporating the calibration techniques demonstrated in \citet{daw2023reds}, \citet{yoo2023using}, and \citet{bonas2024calibrated}. Specifically, we repeatedly fit the proposed model to the dataset using different sets of randomly sampled weights and potentially varying hyperparameters. For each forecast time \( t \) at location \( s \), we generate \( K \) ensemble forecasts, denoted as \( \widehat{Y}^1_{t, s}, \dots, \widehat{Y}^K_{t, s} \). The calibrated prediction interval \( (\widehat{l}_{t, s}, \widehat{u}_{t, s}) \) is then obtained by solving the following optimization problem
\begin{equation}
    (\widehat{l}_{t, s}, \widehat{u}_{t, s}) = \operatorname*{argmin}_{(l_{t, s}, u_{t, s})} (u_{t, s} - l_{t, s}) \label{eq:uncertain},
\end{equation}
subject to the constraint:
\begin{equation*}
\frac{1}{K} \sum_{k=1}^K I(l_{t, s} \leq \widehat{Y}^k_{t, s} \leq u_{t, s}) = 1 - \alpha,
\end{equation*}
where \( \alpha \) is the desired significance level. We note that the prediction interval obtained from this approach corresponds to the unimodal highest density region proposed by \citet{hyndman1996computing}. While we acknowledge that the behavior of ESN forecasts can be quite complex, we choose to use the unimodal highest density region for computational efficiency, as commonly demonstrated in the ESN literature \citep{yoo2023using, daw2023reds, bonas2024calibrated}. Although this unimodal interval is preferred for practical reasons, users may still apply any calibration method after generating the ensemble of forecasts.

Finally, we include the full algorithm for ensemble AESN Algorithm~\ref{alg:aesn}.
\begin{algorithm}[H]
\caption{Ensemble Forecasting with AESN}
\label{alg:aesn}
\begin{algorithmic}[1]
\State \textbf{Input:} $\{\boldsymbol{Y}_t\}$ and $\{\tilde{\boldsymbol{x}}_t\}$ for $t=1, \dots, T$
\State \textbf{Tuning:} Select optimal parameters $\{k, a_u, a_{in}, a_{res}, \gamma^{in}, \gamma^{res}, \alpha, \nu\}$ using random search and empirical knowledge.
\For{$l = 1, \dots, L$}
    \State Simulate $U$ using (\ref{eq:u_embed})
    \State Calculate $\{z_t: t=1:T\}$ using (\ref{eq:ARER})
    \State Simulate $W_{res}$ and $W_{in}$ using (\ref{eq:esn_wres}) and (\ref{eq:esn_win})
    \State Calculate $\{h_t: t=1:T\}$ using (\ref{eq:aesn_h1}) and (\ref{eq:aesn_h2})
    \State Estimate $W_{out}$ using ridge regression
    \State Obtain forecasts $\widehat{\boldsymbol{Y}}_{T+1}^l$
\EndFor
\State Denote the ensemble of forecasts $\widehat{\boldsymbol{Y}}_{T+1}^{ensemble} = \{\widehat{\boldsymbol{Y}}_{T+1}^l: l = 1, \dots L\}$
\State \textbf{Output} ensemble forecast mean $\overline{\boldsymbol{Y}}_{T+1} = \text{mean}(\widehat{\boldsymbol{Y}}_{T+1}^{ensemble})$, ensemble forecast interval $(\widehat{l}_{t, s}, \widehat{u}_{t, s})$ according to (\ref{eq:uncertain})
\end{algorithmic}
\end{algorithm}

\section{Tourism Occupancy Analysis}
\label{sec:analysis}
We demonstrate the performance of the proposed method on Eurostat's tourism occupancy data. Tourism plays a crucial role in shaping regional economies. For instance, infrastructure developed for tourism purposes can significantly boost regional development. Additionally, the creation of numerous job opportunities associated with tourism can enhance the living standards of local residents and contribute to the improvement of local welfare. On the other hand, tourism can pose several challenges to local communities. A large flow of tourists can place considerable pressure on local ecosystems, resulting in significant impacts on local ecology. Without proper management, increased waste, noise, and transportation emissions can lead to serious pollution issues that threaten local wildlife and contribute to habitat loss. Furthermore, the strain on natural resources and public infrastructure, along with overcrowding, can disrupt residents' daily lives and place additional stress on local services. These challenges can be mitigated through careful planning and sustainable management practices that leverage tourism occupancy forecasts to minimize negative impacts while maximizing the benefits of tourism \citep{seghir2015tourism, brida2016has, e2018analysing, doi/10.2785/606702}.

In this section, we aim to forecast future tourism occupancy by analyzing the number of nights spent at tourist accommodations. We compare the performance of our proposed AESN against standard ESN, ESN with EOF, and DLM models for both short-lead (2 months) and long-lead (1 year) forecasts. Additionally, we explore both the spatial and temporal dynamics of tourism occupancy through the models' forecasts.

\subsection{Tourism Occupancy Data}
The variable used to represent tourism occupancy is the number of nights spent at tourist accommodation establishments, sourced from Eurostat \citep{tour-occ-nin2m-2024-9-12, doi/10.2785/606702}. This data is collected through sample surveys or census of various accommodation establishments from business register, and then processed by the competent national statistics authorities \citep{tour-occ-nin2m-2024-9-12, doi/10.2785/606702}. In compliance with Commission Delegated Regulation (EU) 2019/1681 of 1 August 2019, these establishments are required to report monthly data on nights spent at tourist accommodations for NUTS level 2 regions \citep{doi/10.2785/606702}. As defined by the Statistical Classification of Economic Activities in the European Community, these establishments include hotels, holiday and short-stay accommodations, as well as camping grounds, recreational vehicle parks, and trailer parks.

\begin{figure}[H]
\centering
\includegraphics[width=0.8\textwidth]{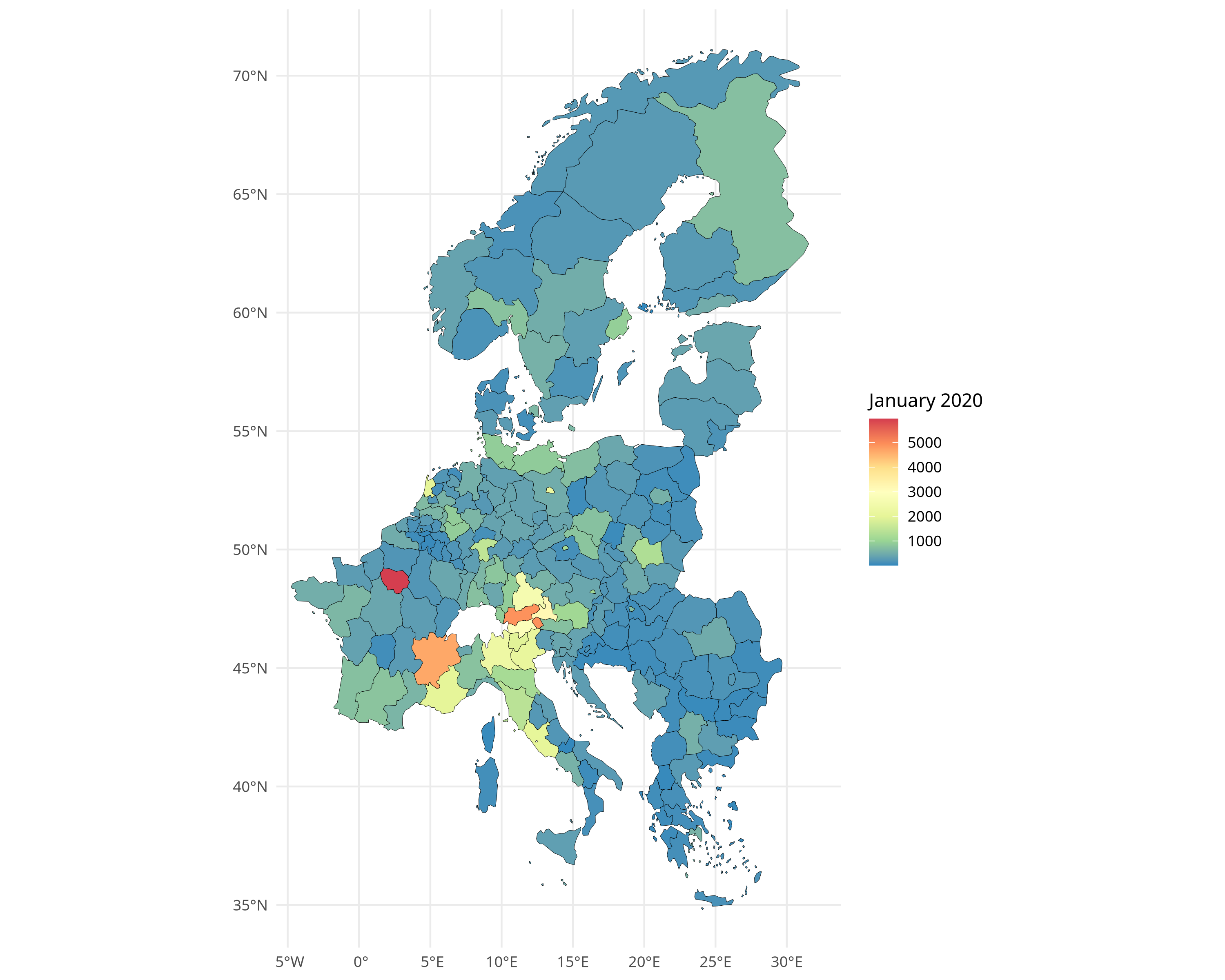}
\caption{Night spent in thousands at tourist accommodations for NUTS level 2 regions in EU on January 2020. The mid-south region tends to attract a higher number of tourists, whereas the eastern and northern regions generally experience lower tourist activity.}
\label{fig:data_spatial}
\end{figure}

The dataset consists of monthly nights spent at both domestic and foreign tourist accommodations from January 2020 to December 2023, covering 242 NUTS level 2 regions in the EU. We excluded regions with missing data at any time point, resulting in a final dataset of 228 locations with 48 time points each. As shown in Figure~\ref{fig:data_spatial},  due to the geographic nature of EU regions, many isolated areas are present, which can be effectively managed by the self-loops in the areal random representation, as discussed in Section~\ref{sec:gcn}. For this study, we use the first 36 months as training data and forecast the last 12 months. In addition, we use the lagged values of the inputs, with the optimal number of lagged values determined together with the hyperparameters. In this analysis, the optimal number of lagged values is 5. The night spent are modeled in thousands on a logarithmic scale and transformed back after forecasting.

Figure~\ref{fig:data_spatial} presents a snapshot of nights spent at tourist accommodations in the EU in January 2020. A clear spatial pattern emerges, with higher numbers of nights spent in regions across mid-southern Europe, particularly in France, Italy, Austria, and Slovenia. These areas show a distinct concentration of tourist activity compared to other parts of Europe.

\begin{figure}[H]
\centering
\includegraphics[width=0.8\textwidth]{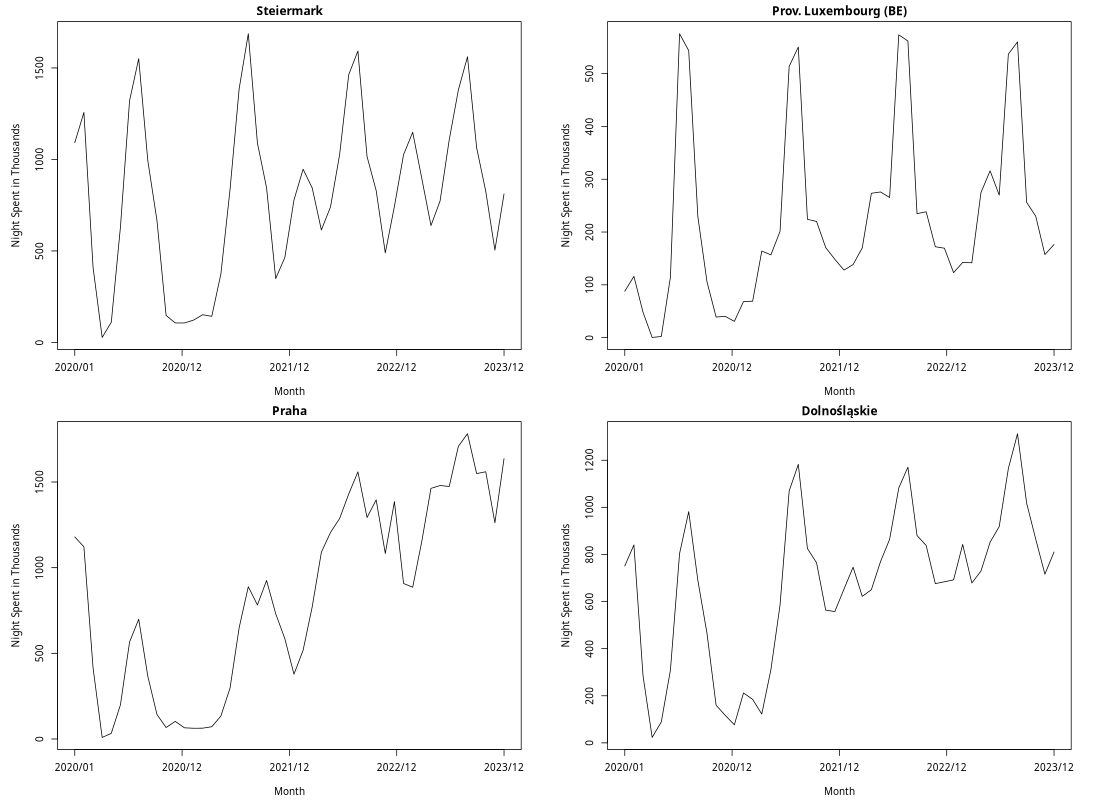}
\caption{Time series plots of night spent in thousands at tourist accommodations at selected locations. All four locations reveal some seasonality, an increasing trend, and other nonlinear characteristics.}
\label{fig:data_time}
\end{figure}

We also demonstrate that the nights spent at tourist accommodations dataset exhibits complex temporal dynamics, as shown in Figure~\ref{fig:data_time}, which presents data from four selected regions. The dataset is influenced by multiple temporal components, with seasonality being one of the most prominent. However, the overall dynamics are intricate, making it difficult to achieve strong forecasting performance using a traditional DLM.

\subsection{Implementation}
\label{sec:implementation}
For ESN-based models, including standard ESN, ESN with EOF, and AESN, we used a validation set approach to select the optimal hyperparameters. Specifically, we further split the training data into 24 months for training and the remaining 12 months as the validation set. We then performed random searches to identify the best-performing hyperparameters on the validation set, as detailed in Section \ref{sec:tune}. Once the optimal parameters were selected, we refit the ESN based models using the entire 36 months in the training data. We used 500 ensembles for all ESN-based models, as this ensures consistent results for our data analysis. We present the relevant optimal hyperparameters for AESN in Table~\ref{tb:params}. Sensitivity analysis shows that AESN tends to favor smaller values of $K$, $n_h$, and $a_u$ for simpler tasks such as the 3-step forecast, while it prefers larger values for more challenging tasks like the 12-step forecast. We also note that due to its randomness, the random search may not always find the most parsimonious configurations. For the graph construction, we use unweighted graph with the Queen adjacency matrix, where diagonal entries are set to zero and non-zero off-diagonal entries indicate neighboring relationships. In practice, users may also consider alternative weighted graphs, such as those based on economic distance. The selection of the number of EOFs follows a randomization approach. Specifically, we compare the singular values of the training data with those of an equivalent uncorrelated dataset, created by randomly rearranging each column of the training data. The optimal number of EOFs is chosen as the point where the traces of the singular values intersect \citep{hastie2009elements,wikle2019spatio}. Using this method, the first 8 EOFs are selected. For comparison, we used the dlm package \citep{dlm_r, dlm_book}. The DLM we used in this study included a second-order polynomial component and a seasonal component with a frequency of 12. The parameters in the state and data models were estimated using the maximum likelihood method provided by the dlm package. We implemented all four models on an AMD Ryzen 5 7640U (12) @ 4.97 GHz CPU with 32 GB of RAM. The DLM model was implemented using the dlm package in R, and all ESN-based models were implemented using NumPy in Python \citep{harris2020array}.
\begin{table}[H]
    \caption{Optimal Parameters for AESN}
    \begin{tabular}{lcccc}
        \hline
        Forecast & $K$ & $a_u$  & $n_h$ & $\nu$ \\
        \hline
        3-step   & 16   & 0.094  & 168   & 0.414 \\
        12-step  & 122  & 0.958  & 253   & 0.072 \\
        \hline
    \end{tabular}
    \label{tb:params}
\end{table}

\subsection{Forecasting Results}
To assess model performance, we evaluate root mean square error (RMSE), continuous ranked probability score (CRPS), and interval score (IS). Notably, CRPS evaluates the forecasting performance over the entire distribution, making it a useful measure of uncertainty quantification. In this study, we employ the empirical form of CRPS as outlined by \citet{zamo2018estimation}.

As shown in Table~\ref{tab:result}, the AESN outperforms all other models in both short-lead (3-step ahead) and long-lead (12-step ahead) forecasts, achieving the lowest RMSE, CRPS, and IS, indicating superior accuracy and uncertainty quantification. ESN with EOF also demonstrates strong performance, particularly with lower errors compared to the standard ESN. In contrast, the DLM struggles significantly in long-lead forecasts, with an extremely high RMSE and IS, making it the least effective model for long-lead forecasting. In terms of runtime, all ESN-based models are clearly more efficient than the DLM model. Among them, the ESN with EOF is the fastest model, which is expected since it only considers the first eight principal components, as discussed in Section~\ref{sec:implementation}. We note that AESN achieves significant improvements in modeling area-level data with only minimal computational cost. In addition, the 3-step forecasts are slower because they require model updates whenever new data becomes available after each 3-step interval. In practice, runtime can also be affected by the choice of hyperparameters. For example, when the analysis requires a small value of $K$, AESN can be as fast as ESN.
\begin{table}[H]
\centering
\caption{Model comparison of short-lead and long-lead forecasts using Eurostat's dataset on nights spent in thousands at tourist accommodations on the original scale.}
\resizebox{\textwidth}{!}{
\begin{tabular}{lcccc cccc}
\toprule
Model & \multicolumn{4}{c}{3-step Forecasts} & \multicolumn{4}{c}{12-step Forecasts} \\
\cmidrule(lr){2-5} \cmidrule(lr){6-9}
                         & RMSE        & CRPS       & IS         & Time    & RMSE         & CRPS       & IS            & Time\\
\midrule
DLM                      & 1927.990    & 243.044    & 4172.548   & 1.57 hours    & 27418371.293 & 251.873    & 1275719.846   & 24.91 mins   \\
ESN                      & 477.667     & 166.814    & 1359.580   & 4.68 mins     & 675.250      & 205.312    & 2391.955      & 1.23 mins   \\
ESN with EOF             & 326.096     & 103.508    & 1135.795   & 3.90 mins     & 477.667      & 166.814    & 1359.580      & 0.93 mins   \\
AESN                     & 202.307     & 60.429     & 561.487    & 5.52 mins     & 332.766      & 91.588     & 908.191       & 2.21 mins   \\
\bottomrule
\end{tabular}
}
\label{tab:result}
\end{table}

\begin{figure}[H]
\centering
\includegraphics[width=0.8\textwidth]{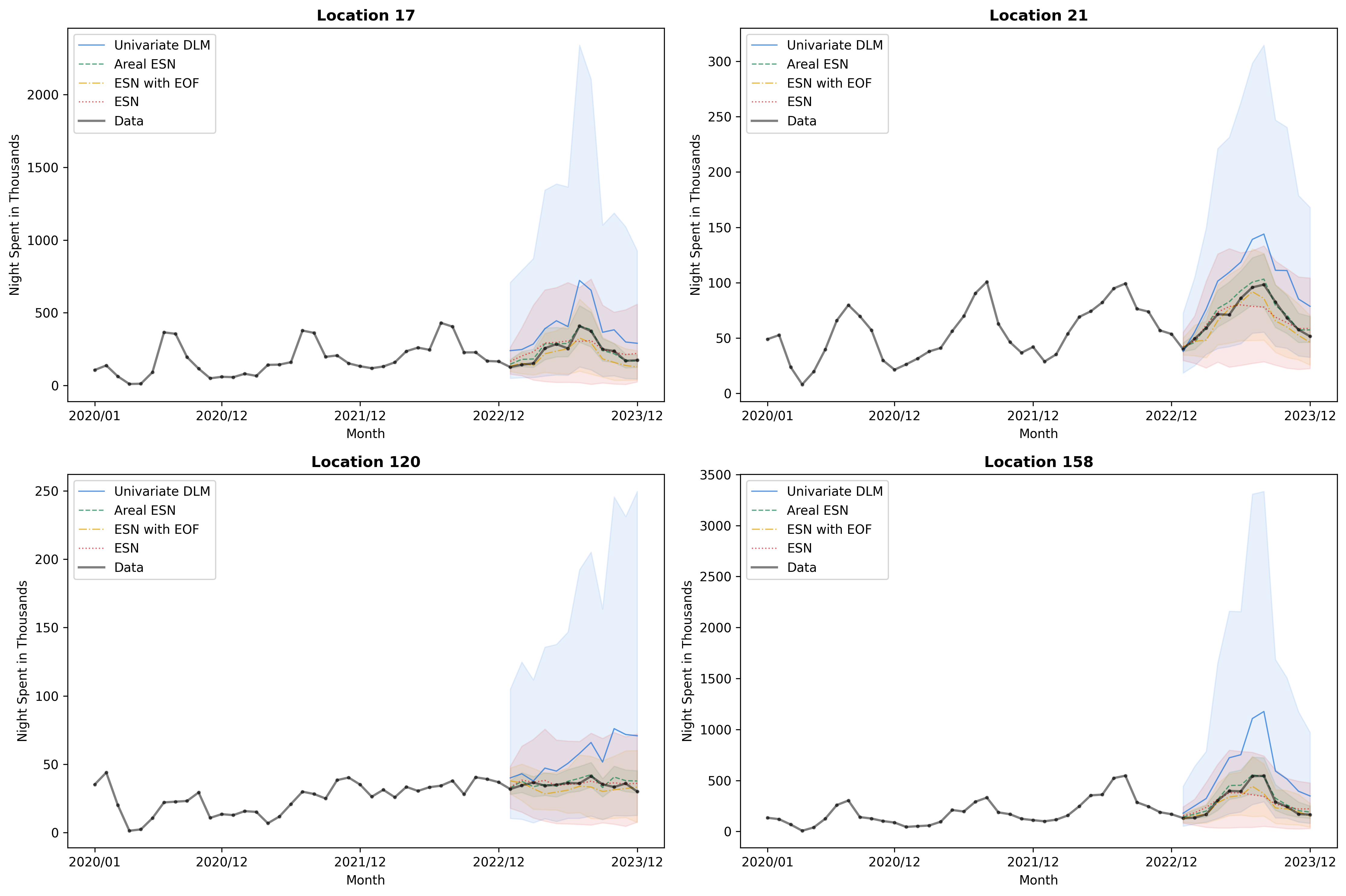}
\caption{Out-of-sample 12-month forecasts for night  spent in thousands at tourist accommodations at four selected locations from January 2023 to December 2023, based on the 12-step forecast of the AESN model. For locations with strong seasonality, the AESN model accurately captures the trend of higher tourism activity during the summer and lower activity in other seasons.}
\label{fig:result_time}
\end{figure}

We demonstrate that the AESN effectively captures the temporal dynamics within the dataset of nights spent at tourist accommodations. This model adeptly handles seasonality and other nonlinear dynamics across various locations, aligning closely with the observed data as illustrated in Figure~\ref{fig:result_time}. In particular, its ability to accurately model the peak tourist seasons, such as summer, where it reflects the high flow of tourists with remarkable precision. In addition, the performance of the AESN is consistently significantly superior to other models across all metrics, as shown in Table~\ref{tab:result}.

\begin{figure}[H]
\centering
\includegraphics[width=0.8\textwidth]{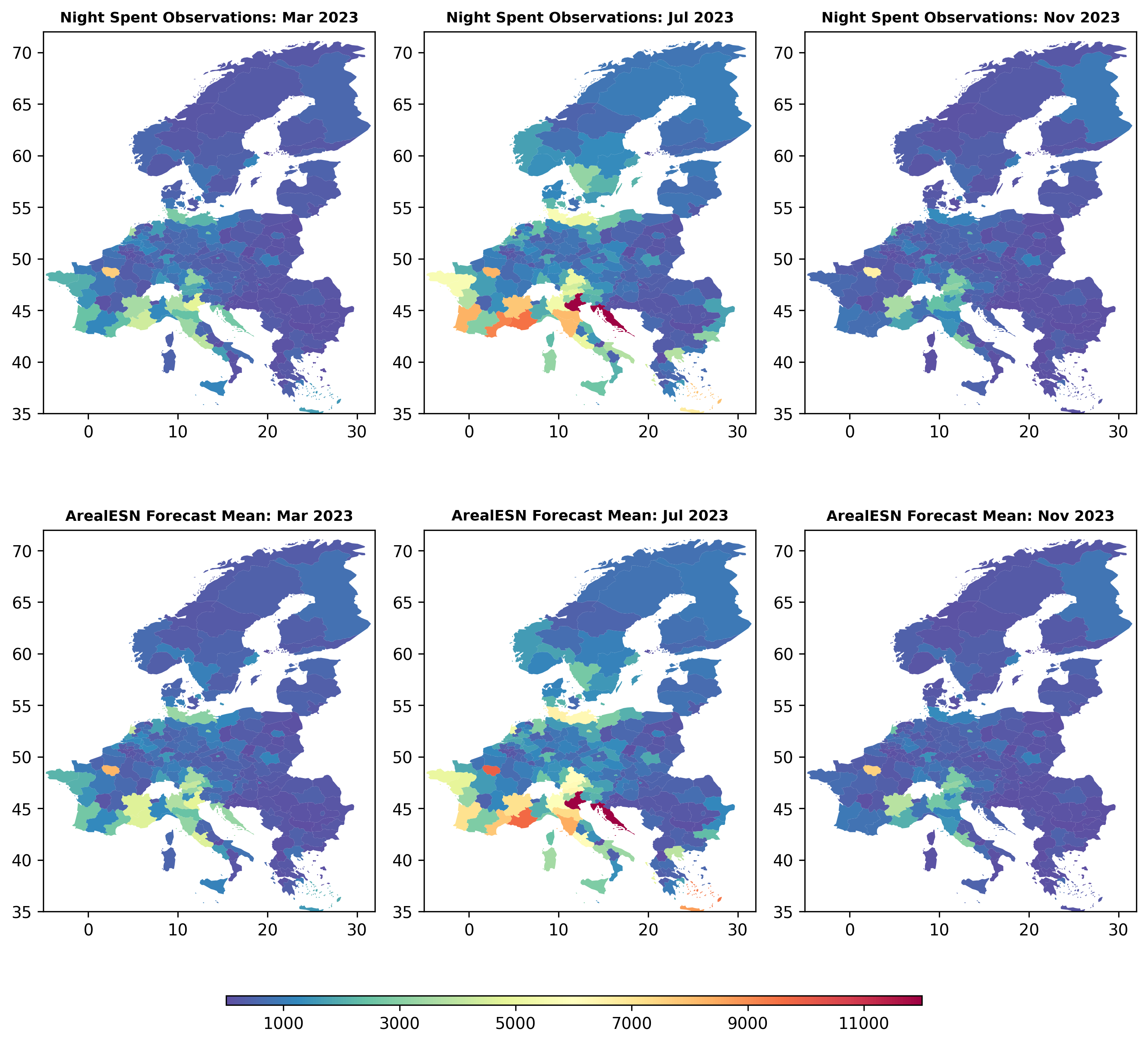}
\caption{Spatial maps of AESN forecasts for nights spent in thousands at tourist accommodations, based on the 12-step ahead forecast of the AESN model. The AESN model effectively captures the spatial pattern of higher tourism in the mid-south regions. Additionally, it accurately reflects the seasonal trend, with tourism activity increasing from low in spring to high in summer and then declining again in winter.}
\label{fig:result_spatial}
\end{figure}

In our analysis, we further explore the AESN's capability in capturing the spatial dynamics across different regions. As depicted in Figure~\ref{fig:result_spatial}, the model shows outstanding performance in accurately mapping the geographical distribution of tourist activity. It effectively highlights the higher tourist influx in the mid-southern regions, contrasting with the less tourism activity in the northern regions. These insights, derived from the 12-step ahead forecasts, can greatly validate the model's precision. 

\section{Discussion}
\label{sec:discussion}
Spatio-temporal area-level datasets are crucial in the realm of official statistics, providing essential insights for policy-making and regional planning. Accurate modeling of these datasets and the generation of reliable forecasts are important for policymakers to make informed strategies for future plannings. These datasets often exhibit complex temporal dynamics and neighborhood structures. While ESNs have been shown to effectively capture nonlinear temporal dynamics, the equally important spatial dynamics have been less explored within this framework. Overlooking these spatial components can greatly reduce the accuracy and usefulness of the forecasts.

In this paper, we introduce a novel approach that combines random representation and GCNs to adeptly handle the spatial structure in ESNs. The efficiency of our method stems from the nature of the random representation, where most parameters are randomly sampled and then fixed, reducing the computational overhead while improving model performance. We demonstrate that our model significantly outperforms traditional DLMs and standard ESN approaches, including those enhanced with EOFs using the Eurostat tourism occupancy dataset. Our results show that the model not only captures the long-lead temporal dynamics effectively but also accurately reflects the intricate spatial dynamics within the data. This capability makes it exceptionally useful for understanding and predicting complex patterns in regional tourism activities, thereby aiding stakeholders in making more precise and effective policies and interventions. Finally, there are several potential directions for future research. First, the current implementation of AESN considers the full graph built on area-level data. Future research should explore training AESN using a mini-batch approach by constructing meaningful subgraphs that account for correlations between subgraphs. Second, missing data is quite common in official statistics due to disclosure limitations and the high costs of conducting surveys at fine geographical levels. Currently, neither the ensemble ESN nor AESN has a mechanism to account for missingness. Incorporating imputation methods could be a valuable direction for future research.

\section*{Acknowledgments}
  This article is released to inform interested parties of ongoing research and to encourage discussion. The views expressed on statistical issues are those of the authors and not those of the NSF or U.S. Census Bureau. This research was partially supported by the U.S. National Science Foundation (NSF) under NSF grant NCSE-2215168.

\bibliography{reference}
\bibliographystyle{jasa}

\end{document}